\def\year{2019}\relax
\documentclass[letterpaper]{article} 
\usepackage{aaai19}  
\usepackage{times}  
\usepackage{helvet}  
\usepackage{courier}  
\usepackage{url}  
\usepackage{graphicx}  
\usepackage{amssymb}
\usepackage{amsmath}
\usepackage{graphicx}

\usepackage{bm}
\newcommand{\vect}[1]{\bm{#1}}
\newcommand{\matr}[1]{\bm{#1}}

\newcommand{\vb}[0]{\vect{b}}
\newcommand{\vc}[0]{\vect{c}}

\newcommand{\vh}[0]{\vect{h}}

\newcommand{\vx}[0]{\vect{x}}
\newcommand{\vz}[0]{\vect{z}}

\newcommand{\vy}[0]{\vect{y}}

\newcommand{\vV}[0]{\vect{V}}
\newcommand{\vr}[0]{\vect{r}}
\newcommand{\vp}[0]{\vect{p}}
\newcommand{\vtheta}[0]{\vect{\theta}}

\newcommand{\mW}[0]{\matr{W}}

\newcommand{\mE}[0]{\matr{E}}

\newcommand{\mU}[0]{\matr{U}}

\newcommand{\mY}{\matr{Y}}
\usepackage{color}
\definecolor{olive}{RGB}{0,184,136}

\begin{document}
%
\title{Bringing back simplicity and lightliness into neural image captioning}
\author{Jean-Benoit Delbrouck \and St\'ephane Dupont \\
         TCTS Lab, University of Mons, Belgium\\
          \{jean-benoit.delbrouck, stephane.dupont\}@umons.ac.be}

\maketitle
\begin{abstract}
 Neural Image Captioning (NIC) or neural caption generation has attracted a lot of attention over the last few years. Describing an image with a natural language has been an emerging challenge in both fields of computer vision and language processing. Therefore a lot of research has focused on driving this task forward with new creative ideas. So far, the goal has been to maximize scores on automated metric and to do so, one has to come up with a plurality of new modules and techniques. Once these add up, the models become complex and resource-hungry. In this paper, we take a small step backwards in order to study an architecture with interesting trade-off between performance and computational complexity. To do so, we tackle every component of a neural captioning model and propose one or more solution that lightens the model overall. Our ideas are inspired by two related tasks: Multimodal and Monomodal Neural Machine Translation.
\end{abstract}

\section{Introduction} \label{introduction}

Problems combining vision and natural language processing such as image captioning \cite{DBLP:journals/corr/ChenFLVGDZ15} is viewed as an extremely challenging task. It requires to grasp and express low to high-level aspects of local and global areas in an image as well as their relationships. Over the years, it continues to inspire considerable research. Visual attention-based neural decoder models \cite{pmlr-v37-xuc15,conf/cvpr/KarpathyL15} have shown gigantic success and are now widely adopted for the NIC task. These recent advances are inspired from the neural encoder-decoder framework \cite{SutskeverVL14,bahdanau+al-2014-nmt}---or sequence to sequence model (seq2seq)--- used for Neural Machine Translation (NMT). In that approach, Recurrent Neural Networks (RNN, \citeauthor{conf/interspeech/MikolovKBCK10} \citeyear{conf/interspeech/MikolovKBCK10}) map a source sequence of words (encoder) to a target sequence (decoder). An attention mechanism is learned to focus on different parts of the source sentence while decoding. The same mechanism applies for a visual input; the attention module learns to attend the salient parts of an image while decoding the caption. \\

These two fields, NIC and NMT, led to a Multimodal Neural Machine Translation (MNMT, \citeauthor{specia-EtAl:2016:WMT} \citeyear{specia-EtAl:2016:WMT}) task where the sentence to be translated is supported by the information from an image. Interestingly, NIC and MNMT share a very similar decoder: they are both required to generate a  meaningful natural language description or translation with the help of a visual input. However, both tasks differ in the amount of annotated data. MNMT has $\approx$ 19 times less unique training examples, reducing the amount of learnable parameters and potential complexity of a model. Yet, over the years, the challenge has brought up very clever and elegant ideas that could be transfered to the NIC task. The aim of this paper is to propose such an architecture for NIC in a straightforward manner. Indeed, our proposed models work with less data, less parameters and require less computation time. More precisely, this paper intents to:

\begin{figure}
\centering
\includegraphics[scale=0.75]{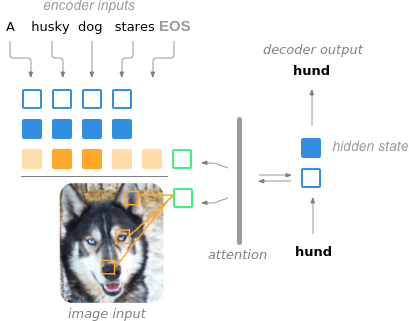}
\caption{This image depicts a decoder timestep of the MNMT architecture. At time $t$, the decoder attend both a visual and textual representations. In NIC, the decoder only attends an image. This shows how both tasks are related.}\label{fig}
\end{figure}

\begin{itemize}
\item Work only with in-domain data. No additional data besides proposed captioning datasets are involved in the learning process;
\item Lighten as much as possible the training data used, i.e. the visual and linguistic inputs of the model;
\item Propose a subjectively light and straightforward yet efficient NIC architecture with high training speed.
\end{itemize}

\section{Captioning Model}
As quickly mentionned in section~\ref{introduction}, a neural captioning model is a RNN-decoder~\cite{bahdanau+al-2014-nmt} that uses an attention mechanism over an image $I$ to generate a word $y_t$ of the caption at each time-step $t$. The following equations depict what a baseline time-step $t$ looks like~\cite{pmlr-v37-xuc15}:

\begin{align}
\vx_t &= \mW^x \mE y_{t-1}  \label{eq:0} \\ 
\vc_t &= f_{\text{att}}(\vh_{t-1}, I) \label{eq:1}\\ 
\vh_t &= f_{\text{rnn}}(\vx_t, \vh_{t-1}, \vc_t)  \label{eq:2} \\
y_t \sim \vp_t &= \mW^y [\vy_{t-1}, \vh_{t},\vc_t] \label{eq:3}
\end{align}

where equation~\ref{eq:0} maps the previous embedded word generated to the RNN hidden state size with matrice $\mW^x$ , equation~\ref{eq:1} is the attention module over the image $I$, equation~\ref{eq:2} is the RNN cell computation and equation~\ref{eq:3} is the probability distribution $\vp_t$ over the vocabulary (matrix $\mW^y $ is also called the  projection matrix). 

If we denote $\vtheta$ as the model parameters, then $\vtheta$ is learned by maximizing the likelihood of the observed sequence $\mY = (\vy_1, \vy_2, \cdots, \vy_n)$ or in other words by minimizing the cross entropy loss. The objective function is given by:
\begin{equation}
\mathcal{L}(\vtheta) = - \sum\limits_{t=1}^n \log p_{\vtheta}(\vy_t | \vy<t, I) \label{eq:5}
\end{equation}

The paper is structured so that each section tackles an equation (i.e. a main component of the captioning model) in the following manner: section \ref{embed} for equation \ref{eq:0} (embeddings), section \ref{cgrusec} for equation \ref{eq:2} ($f_{\text{rnn}}$), section \ref{attmodel} for equation \ref{eq:1} ($f_{\text{att}}$), section \ref{projection} for equation \ref{eq:3} (projection) and section \ref{reinf} for equation \ref{eq:5} (objective function).

\subsection{Embeddings} \label{embed}
\null\hfill $^{\text{Recall that: }}\vx_t = \mW^x \mE y_{t-1}$ \\ 

The total size of the embeddings matrix $\mE$ depends on the vocabulary size $|\mathcal{Y}|$ and the embedding dimension $d$ such that $\mE \in \mathbb{R}^{|\mathcal{Y}|\times d}$. The mapping matrix $\mW^x$ also depends on the embedding dimension  because $\mW^x \in \mathbb{R}^{d \times |\vx_t|}$. \\

Many previous researches \cite{conf/cvpr/KarpathyL15,You2016ImageCW,yao2017boosting,Anderson2017up-down} uses pretrained embeddings such as Glove and word2vec or one-hot-vectors. Both word2vec and glove provide distributed representation of words. These models are pre-trained on 30 and 42 billions words respectively \cite{NIPS2013_5021,pennington2014glove}, weights several gigabyes and work with $d = 300$. \\

For our experiments, each word $y_i$ is a column index in an embedding matrix $\mE_y$ learned along with the model and initialized using some random distribution. Whilst the usual allocation is from 512 to 1024 dimensions per embedding \cite{pmlr-v37-xuc15,Lu2017Adaptive,mun2017textguided,Rennie2017SelfCriticalST}  we show that a small embedding size of $d = 128$ is sufficient to learn a strong vocabulary representation. The solution of an jointly-learned embedding matrix also tackles the high-dimensionality and sparsity problem of one-hot vectors. For example, \cite{Anderson2017up-down} works with a vocabulary of 10,010 and a hidden size of 1000. As a result, the mapping matrix $\mW^x$ of equation \ref{eq:0} has 10 millions parameters.\\

Working with a small vocabulary, besides reducing the size of embedding matrix $\mE$, presents two major advantages: it lightens the projection module (as explained further in section \ref{projection}) and reduces the action space in a Reinforcement Learning setup (detailed in section \ref{reinf}). To marginally reduce our vocabulary size (of $\approx$ 50 \%), we use the byte pair encoding (BPE) algorithm on the train set to convert space-separated tokens into subwords \cite{P16-1162}. Applied originally for NMT, BPE is based on the intuition that various word classes are made of smaller units than words such as compounds and loanwords. In addition of making the vocabulary smaller and the sentences length shorter, the subword model is able to productively generate new words that were not seen at training time.

\subsection{Conditional GRU} \label{cgrusec}

\null\hfill $^{\text{Recall that: }}\vh_t = f_{\text{rnn}}(\vx_t, \vh_{t-1}, \vc_t)$ \\

Most previous researches in captioning~\cite{conf/cvpr/KarpathyL15,You2016ImageCW,Rennie2017SelfCriticalST,mun2017textguided,Lu2017Adaptive,yao2017boosting} used an LSTM~\cite{Hochreiter:1997:LSM:1246443.1246450} for their $f_{\text{rnn}}$ function. Our recurrent model is a pair of two Gated Recurrent Units (GRU \cite{cho-al-emnlp14}), called conditional GRU (cGRU), as previously investigated in NMT\footnote{\url{https://github.com/nyu-dl/dl4mt-tutorial/blob/master/docs/cgru.pdf}}. A GRU is a lighter gating mechanism than LSTM since it doesn't use a forget gate and lead to similar results in our experiments. \\

The cGRU also addresses the encoding problem of the $f_{\text{att}}$ mechanism. As shown in equation~\ref{eq:1} the context vector $\vc_t$ takes the previous hidden state $\vh_{t-1}$ as input which is outside information of the current time-step. This could be tackled by using the current hidden $\vh_{t}$, but then context vector $\vc_t$ is not an input of $f_{\text{rnn}}$ anymore. A conditional GRU is an efficient way to both build and encode the result of the $f_{\text{att}}$ module. \\

Mathematically, a first independent GRU encodes an intermediate hidden state proposal $\vh_t^{\prime}$ based on the previous hidden state $\vh_{t-1}$ and input $\vx_t$ at each time-step $t$:

\begin{align}\label{cgrublock0}
	\vz_t^{\prime} =& ~ \sigma \left(  \vx_t  + \mU_z^{\prime} \vh_{t-1}  \right) \nonumber \\
	\vr_t^{\prime} =& ~ \sigma \left(  \vx_t + \mU_r^{\prime} \vh_{t-1}  \right)  \nonumber \\ 
	\underline{\vh}_t^{\prime} =& ~\text{tanh} \left( \vx_t + \vr_t^{\prime} \odot (\mU^{\prime}\vh_{t-1})  \right)  \nonumber \\        
	\vh_t^{\prime} =& ~ (1 - \vz_t^{\prime}) \odot \underline{\vh}_t^{\prime} + \vz_t^{\prime} \odot \vh_{t-1}
	\end{align}
    
	Then, the attention mechanism computes $\vc_t$ over the source sentence using the image $I$ and the intermediate hidden state proposal $\vh_t^{\prime}$ similar to \ref{eq:2}:
	$$\vc_t = f_{\text{att}}(\vh_t^{\prime}, I) $$
    
   	Finally, a second independent GRU computes the hidden state $\vh_t$ of the $\text{cGRU}$ by looking at the intermediate representation $\vh_t^{\prime}$ and context vector $\vc_t$:
    
	\begin{align} \label{cgrublock}
	\vz_t =& ~ \sigma \left( \mW_z \vc_t + \mU_z \vh_t^{\prime} \right) \nonumber \\
	\vr_t =& ~ \sigma \left( \mW_r \vc_t + \mU_r \vh_t^{\prime} \right) \nonumber \\
	\underline{\vh}_t =& ~ \text{tanh} \left(  \mW \vc_t  + \vr_t \odot (\mU \vh_t^{\prime} )  \right)  \nonumber \\       
	\vh_t =& ~ (1 - \vz_t) \odot \underline{\vh}_t + \vz_t \odot \vh_t^{\prime}
	\end{align}	
    
We see that both problem are addressed: context vector $\vc_t$ is computed according to the intermediate representation $\vh_t^{\prime}$ and the final hidden state $\vh_t$ is computed according to the context vector $\vc_t$. Again, the size of the hidden state $|\vh_t|$ in the literature varies between 512 and 1024, we pick $|\vh_t|$ = 256.

The most similar approach to ours is the Top-Down Attention of \cite{Anderson2017up-down} that encodes the context vector the same way but with LSTM and a different hidden state layout.

\subsection{Attention model} \label{attmodel}
\null\hfill $^{\text{Recall that: }}\vc_t = f_{\text{att}}(\vh_{t-1}, I)$ \\

Since the image is the only input to a captioning model, the attention module is crucial but also very diverse amongst different researches. For example, \cite{You2016ImageCW} use a semantic attention where, in addition of image features, they run a set of attribute detectors to get a list of visual attributes or concepts that are most likely to appear in the image. \cite{Anderson2017up-down} uses the Visual Genome dataset to pre-train his bottom-up attention model. This dataset contains 47,000 out-of-domain images of the capioning dataset densely annotated with scene graphs containing objects, attribute and relationships. \cite{YangReview} proposes a review network which is an extension to the decoder. The review network performs a given number of review steps on the hidden states and outputs a compact vector representation available for the attention mechanism. \\
 
Yet everyone seems to agree on using a Convolutional Neural Network (CNN) to extract features of the image $I$. The trend is to select features matrices, at the convolutional layers, of size $14 \times 14 \times 1024$ (Resnet, \citeauthor{He2016DeepRL} \citeyear{He2016DeepRL}, res4f layer) or $14 \times 14 \times 512$ (VGGNet \citeauthor{Simonyan2014VeryDC} \citeyear{Simonyan2014VeryDC}, conv5 layer). Other attributes can be extracted in the last fully connected layer of a CNN and has shown to bring useful information \cite{YangReview,yao2017boosting,You2016ImageCW} Some models also finetune the CNN during training \cite{YangReview,mun2017textguided,Lu2017Adaptive} stacking even more trainable parameters. \\

Our attention model $f_{\text{att}}$ is guided by a unique vector with global 2048-dimensional visual representation $\vV$ of image $I$ extracted at the pool5 layers of a ResNet-50. Our attention vector is computed so:

\begin{equation}
\vc_t = \vh_t^{\prime} \odot \text{tanh} (\mW^{\text{img}} \vV_I)
\end{equation}

Recall that following the cGRU presented in section \ref{cgrusec}, we work with $\vh_t^{\prime}$ and not $\vh_{t-1}$. Even though pooled features have less information than convolutional features ($\approx$ 50 to 100 times less), pooled features have shown great success in combination with cGRU in MNMT \cite{W17-4746}. Hence, our attention model is only the single matrice $\mW^{\text{img}} \in \mathbb{R}^{2048 \times |\vh_t^{\prime}|}$

\subsection{Projection} \label{projection}
\null\hfill $^{\text{Recall that: }}y_t \sim \vp_t = \mW^y [\vy_{t-1}, \vh_{t},\vc_t]$ \\

The projection also accounts for a lot of trainable parameters in the captioning model, especially if the vocabulary is large. Indeed, in equation \ref{eq:3} the projection matrix is $\in \mathbb{R}^{[\vy_{t-1}, \vh_{t},\vc_t] \times |\mathcal{Y|}}$. To reduces the number of parameters, we use a bottleneck function:

\begin{align}
\vb_t &= f_{\text{bot}}(\vy_{t-1}, \vh_{t},\vc_t) = \mW^{\text{bot}}[\vy_{t-1}, \vh_{t},\vc_t] \label{f_bt} \\
y_t \sim \vp_t &= \mW^{y\_bot} \vb_t 
\end{align}

where $|\vb_t| < |[\vy_{t-1}, \vh_{t},\vc_t]|$ so that $|\mW^{bot} | + |\mW^{y\_bot} | < ||\mW^{y}|$. Interestingly enough, if $|\vb_t| = d$ (embedding size), then $|\mW^{y\_bot}| = |\mE|$. We can share the weights between the two matrices (i.e. $\mW^{y\_bot} = \mE$) to marginally reduce the number of learned parameters. Moreover, doing so doesn't negatively impact the captioning results.

We push our projection further and use a deep-GRU, used originally in MNMT \cite{delbrouck2018umons}, so that our bottleneck function $f_{\text{bot}}$ is now a third GRU as described by equations \ref{cgrublock}:

\begin{equation}
\vb_t = f_{\text{bot}}(\vy_{t-1}, \vh_t^{\prime},\vc_t) = \text{cGRU}([\vy_{t-1}, \vh_t^{\prime},\vc_t], \vh_t)  \\
\end{equation}

Because we work with small dimension, adding a new GRU block on top barely increases the model size.

\subsection{Objective function} \label{reinf}.

To directly optimize a automated metric, we can see the captioning generator as a Reinforcement Learning (RL) problem. The introduced $f_{\text{rnn}}$ function is viewed as an agent that interact with an environment composed of words and image features. The agent interacts with the environment by taking actions that are the prediction of the next word of the caption. An action is the result of the policy $p_{\vtheta}$ where $\vtheta$ are the parameters of the network. Whilst very effective to boost the automatic metric scores, porting the captioning problem into a RL setup significantly reduce the training speed. \\

\citeauthor{DBLP:journals/corr/RanzatoCAZ15} \citeyear{DBLP:journals/corr/RanzatoCAZ15} proposed a method (MIXER), based on the REINFORCE method, combined with a baseline reward estimator. However, they implicitly assume each intermediate action (word) in a partial sequence has the same reward as the sequence-level reward, which is not true in general. To compensate for this, they introduce a form of training that mixes together the MLE objective and the REINFORCE objective. \citeauthor{Liu2017ImprovedIC} \citeyear{Liu2017ImprovedIC} also addresses the delayed reward problem by estimating at each time-step the future rewards based on Monte Carlo rollouts. \citeauthor{Rennie2017SelfCriticalST} \citeyear{Rennie2017SelfCriticalST} utilizes the output of its own test-time inference model to normalize the rewards it experiences. Only samples from the model that outperform the current test-time system are given positive weight. \\

To keep it simple, and because our reduced vocabulary allows us to do so, we follow the work of \citeauthor{DBLP:journals/corr/RanzatoCAZ15} \citeyear{DBLP:journals/corr/RanzatoCAZ15}  and use the naive variant of the policy gradient with REINFORCE. The loss function in equation \ref{eq:5} is now given by:

\begin{equation}
\mathcal{L}(\vtheta) = - \mathbb{E}_{\mY\sim p_{\vtheta}} \vr(\mY) 
\end{equation}

where $\vr(\mY)$ is the reward (here the score given by an automatic metric scorer) of the outputted caption $\mY = (\vy_1, \vy_2, \cdots, \vy_n)$.\\

We use the REINFORCE algorithm based on the observation that the expected gradient of a non-differentiable reward function is computed as follows:

\begin{equation}
\nabla_{\vtheta} \mathcal{L}(\vtheta) = - \mathbb{E}_{\mY\sim p_{\vtheta}} [ \vr(\mY) \nabla_{\vtheta} \log_{p_{\vtheta}}(\mY)]
\end{equation}

The expected gradient can be approximated using $N$ Monte-Carlo sample $\mY$ for each training example in the batch:

\begin{equation}
\nabla_{\vtheta} \mathcal{L}(\vtheta) = \nabla - \bigg[  \frac{1}{N} \sum_{i=1}^{N} [ r_i(\mY_i)  \log_{p_{\vtheta}}(\mY_i)] \bigg]
\end{equation}

In practice, we can approximate with one sample:

\begin{equation}
\nabla_{\vtheta} \mathcal{L}(\vtheta) \approx - \vr(\mY) \nabla_{\vtheta} \log_{p_{\vtheta}}(\mY)
\end{equation}

The policy gradient can be generalized to compute the reward associated with an action value relative to a baseline $\vb$. This baseline either encourages a word choice $\vy_t$ if $r_t > b_t$ or discourages it $r_t < b_t$.  If the baseline is an arbitrary function that does
not depend on the actions $\vy_1, \vy_2, \cdots, \vy_n \in \mY$ then baseline does not change the expected gradient, and importantly, reduces the variance of the gradient estimate. The final expression is given by: 

\begin{equation}
\nabla_{\vtheta} \mathcal{L}(\vtheta) \approx - (\vr(\mY) - \vb) \nabla_{\vtheta} \log_{p_{\vtheta}}(\mY)
\end{equation}

\section{Settings} \label{settings}

Our decoder is a cGRU where each GRU is of size $|\vh_t|$ = 256. Word embedding matrix $\mE$ allocates $d=128$ features per word To create the image annotations used by our decoder, we used a ResNet-50 and extracted the features of size 1024 at the pool-5 layer. As regularization method, we apply dropout with a probability of 0.5 on bottleneck $\vb_t$ and we early stop the training if the validation set CIDER metric does not improve for 10 epochs. All variants of our models are trained with ADAM optimizer \cite{kingma2014adam} with a learning rate of $4e^{-4}$ and mini-batch size of 256. We decode with a beam-search of size 3. In th RL setting, the baseline is a linear projection of $\vh_t$. \\

We evaluate our models on MSCOCO \cite{mscoco}, the most popular benchmark for image captioning which contains 82,783 training images and 40,504 validation images. There are 5 human-annotated descriptions per image. As the annotations of the official testing set are not publicly available, we follow the settings in prior work (or "Kaparthy splits" \footnote{https://github.com/karpathy/neuraltalk2/tree/master/coco}) that takes 82,783 images for training, 5,000 for validation and 5,000 for testing. On the training captions, we use the byte pair encoding algorithm on the train set to convert space-separated tokens into subwords (\citeauthor{P16-1162} \citeyear{P16-1162}, 5000 symbols), reducing our vocabulary size to 5066 english tokens. For the online-evaluation, all images are used for training except for the validation set.

\section{Results} \label{results}

Our models performance are evaluated according to the following automated metrics: BLEU-4 \cite{Papineni:2002:BMA:1073083.1073135}, METEOR \cite{Vedantam_2015_CVPR} and CIDER-D \cite{Lavie:2007:MAM:1626355.1626389}.
Results shown in table \ref{scores} are using cross-entropy (XE) loss (cfr. equation \ref{eq:5}). Reinforced learning optimization results are compared in table \ref{rl-scores}. \\

\subsection{XE scores}

\begin{table*}
\caption{Table sorted per CIDER-D score of models being optimized with cross-entropy loss only (cfr. equation \ref{eq:5}).\\
$\circ$ pool features, $\bullet$ conv features, $\star$ FC features, $\mathsection$ means glove or word2vec embeddings, $\dagger$ CNN finetuning in-domain, $\ddagger$ using in-domain CNN, $\amalg$ CNN finetuning OOD  \\}
\begin{tabular}{lccccccc}
			\multicolumn{1}{c}{ }  & \multicolumn{1}{c}{\bf B4} & \multicolumn{1}{c}{\bf M}
            & \multicolumn{1}{c}{\bf C} & \multicolumn{1}{c}{\bf Wt. (in M)} & \multicolumn{1}{c}{\bf Att. feat. (in K)}
            & \multicolumn{1}{c}{\bf O.O.D. (in M)} & \multicolumn{1}{c}{\bf epoch}
			\\ \hline \\
            \textit{This work} \\
            cGRU $\circ$ & 0.302&0.258&1.018&2.46&2&- & 9\\
            			\\ \hline \\
\textit{Comparable work} \\

Adaptive\cite{Lu2017Adaptive} $\bullet\dagger$ & 0.332 & 0.266 & 1.085 & 17.4 & 100 &-&42\\ 
Top-down \cite{Anderson2017up-down} $\bullet\amalg$ & 0.334 & 0.261 & 1.054 & $\approx$ 25 & 204 \\
Boosting            \cite{yao2017boosting} $\circ\star$ & 0.325 & 0.251& 0.986 & $ \approx$ 28.4 & 2 & - & 123\\ 
Review			\cite{YangReview} $\bullet\star\dagger$& 0.290 & 0.237 &  0.886 & $\approx$ 12.3 & 101 & - & 100\\             

SAT     		\cite{pmlr-v37-xuc15} $\bullet$ & 0.250& 0.230& - & $\approx$ 18 & 100&-&-\\          
 \\ \hline \\
\textit{O.O.D work} \\
Top-down \cite{Anderson2017up-down} $\bullet\amalg$ & 0.362 & 0.27 & 1.135 & $\approx$ 25  & 920 & 920$_{RCNN}$ \\

T-G att           (Mun et al. \citeyear{mun2017textguided}) $\bullet\dagger$ & 0.326& 0.257& 1.024 & $\approx$ 12.8 & 200 & 14$_{\text{TSV}}$ &- \\
 Semantic           \cite{You2016ImageCW} $\circ\star\mathsection$ & 0.304 & 0.243 & - & 5.5& 2 & 3$_{\text{glove}}$ & -\\ 
 NT\cite{conf/cvpr/KarpathyL15}$\bullet\mathsection$& 0.230 & 0.195&  0.660 &-&-&3$_{\text{glove}}$&-\\
\end{tabular}
 \label{scores}
\end{table*}

We sort the different works in table \ref{scores} by CIDER score. For every of them, we detail the trainable weights involved in the learning process (Wt.), the number of visual features used for the attention module (Att. Feat), the amount of out-of-domain data (O.O.D) and the convergence speed (epoch). \\

As we see, our model has the third best METEOR and CIDER scores across the board. Yet our BLEU metric is quiet low, we postulate two potential causes. Either our model has not enough parameters to learn the correct precision for a set of n-grams as the metric would require or it is a direct drawback from using subwords. Nevertheless, the CIDER and METEOR metric show that the main concepts are presents our captions. Our models are also the lightest in regards to trainable parameters and attention features number. As far as convergence in epochs were reported in previous works, our cGRU model is by far the fastest to train.\\

The following table \ref{online} concerns the online evaluation of the official MSCOCO test-set 2014 split. Scores of our model are an ensemble of 5 runs with different initialization.

\setlength{\tabcolsep}{4pt}
\begin{table}[!ht]
\caption{Published Ranking image captioning results on the online MSCOCO test server\\\hspace{\textwidth}}
\begin{tabular}{lccc}
			\multicolumn{1}{c}{ }  & \multicolumn{1}{c}{\bf B4(c5)} & \multicolumn{1}{c}{\textbf{M (c5)}}  & \multicolumn{1}{c}{\bf C(c5)}
            \\ \hline \\
            cGru & 0.326 & 0.253 & 0.973 \\
			\\ \hline \\
 Comparable work \\
\cite{Lu2017Adaptive} &0.336&0.264&1.042 \\
\cite{yao2017boosting}  & 0.330&0.256&0.984 \\
\cite{YangReview} & 0.313 & 0.256 & 0.965 \\
\cite{You2016ImageCW} & 0.316 & 0.250 & 0.943 \\
\cite{Wu_2016_CVPR} & 0.306 & 0.246 & 0.911 \\
\cite{pmlr-v37-xuc15} & 0.277 & 0.251 & 0.865
\end{tabular}
\label{online}
\end{table}

We see that our model suffers a minor setback on this test-set, especially in term of CIDER score whilst the adaptive \cite{Lu2017Adaptive} and boosting method \cite{yao2017boosting} yields to stable results for both test-sets.

\subsection{RL scores}

The table \ref{rl-scores} depicts the different papers using direct metric optimization. \citeauthor{Rennie2017SelfCriticalST} \citeyear{Rennie2017SelfCriticalST} used the SCST method, the most effective one according to the metrics boost (+23, +3 an +123 points respectively) but also the most sophisticated. \citeauthor{Liu2017ImprovedIC} \citeyear{Liu2017ImprovedIC} used a similar approach than ours (MIXER) but with Monte-Carlo roll-outs (i.e. sampling until the end at every time-step $t$). Without using this technique, two of our metrics improvement (METEOR and CIDER) surpasses the MC roll-outs variant (+0 against -2 and +63 against +48 respectively).

\setlength{\tabcolsep}{4pt}
\begin{table}[!ht] 
\caption{All optimization are on the CIDER metric. \\\hspace{\textwidth}}
\begin{tabular}{lcccccc}
			\multicolumn{1}{c}{ }  & \multicolumn{1}{c}{\bf B4} & \multicolumn{1}{c}{}  & \multicolumn{1}{c}{\bf M} & \multicolumn{1}{c}{} & \multicolumn{1}{c}{\bf C} & \multicolumn{1}{c}{} 
			\\ \hline \\
Renn. et al. \citeyear{Rennie2017SelfCriticalST} \\
XE  & 0.296& & 0.252 & &0.940& \\
RL-SCST &0.319& \textcolor{olive}{$\uparrow$ 23}& 0.255 &\textcolor{olive}{$\uparrow$ 3}&1.063&\textcolor{olive}{$\uparrow$ 123}\\
\citeauthor{Liu2017ImprovedIC} \citeyear{Liu2017ImprovedIC} \\
XE  & 0.294& & 0.251& &0.947& \\
RL-PG &0.333& \textcolor{olive}{$\uparrow$ 39}& 0.249 &\textcolor{red}{$\downarrow$ 2}&0.995&\textcolor{olive}{$\uparrow$ 48} \\
Ours \\
XE  & 0.302& &0.258& &1.018& \\
RL-PG &0.315& \textcolor{olive}{$\uparrow$ 13}& 0.258 &&1.071&\textcolor{olive}{$\uparrow$ 63} \\

\end{tabular}
\label{rl-scores}
\end{table}

\subsection{Scalability}
An interesting investigation would be to leverage the architecture with more parameters to see how it scales. We showed our model performs well with few parameters, but we would like to show that it could be used as a base for more complex posterior researches.  \\

We propose two variants to effectively do so :

\begin{itemize}
\item \textit{\textbf{cGRUx2}} \quad The first intuition is to double the width of the model, i.e. the embedding size $d$ and the hidden state size $| \vh_t|$. Unfortunately, this setup is not ideal with a deep-GRU because the recurrent matrices of equations \ref{cgrublock0} and \ref{cgrublock} for the bottleneck GRU gets large. We can still use the classic bottleneck function (equation \ref{f_bt}).
\item \textit{\textbf{MHA}} \quad We trade our attention model described in section \ref{attmodel} for a standard multi-head attention (MHA) to see how convolutional features could improve our CIDER-D metrics.
Multi-head attention \cite{NIPS2017_7181} computes a weighted sum
of some values, where the weight assigned to each value is computed by a compatibility function of the query with the corresponding key. This process is repeated $n$ multiple times. The compatibility function is given by:

$$\text{Attention}(Q, K, V) = \text{softmax}(\frac{QK^T}{\sqrt[]{d_q}})V$$

where the query $Q$ is $\vh_t^{\prime}$, the keys and values are a set of 196 vectors of dimension 1024 (from the layer $\text{res4f\_relu}$ of ResNet-50 CNN). Authors found it beneficial to linearly project the queries, keys and values $n$ times with different learned linear projections to dimension $d_q$. The output of the multi-head attention is the concatenation of the $n$ number of $d_q$ values lineary projected to $d_q$ again. We pick $d_q$ = $|\vh_t^{\prime}|$ and $n=3$. The multi-head attention model adds up 0.92M parameters if $|\vh_t^{\prime}| = 256$ and 2.63M parameters if $|\vh_t^{\prime}| = 512$ (in the case of cGRUx2).
\end{itemize}

\begin{figure}[!ht]
\centering
\includegraphics[scale=0.48]{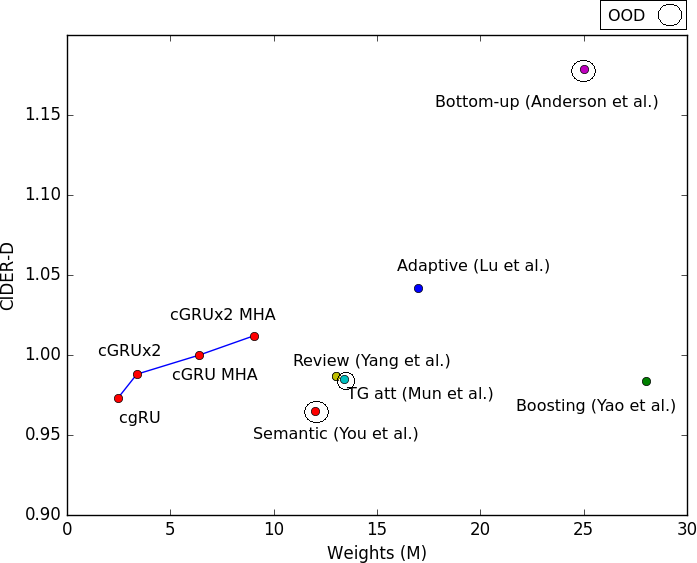}
\caption{The figure shows a new set of results on the online MSCOCO test server and tries to put those in perspective}
\end{figure}

We have hence proposed an image captioning architecture that, compared to previous work, cover a different area in the performance-complexity trade-off plane. We hope these will be of interest and will fuel more research in this direction.

\section{Related work}
As mentioned in the introduction, our model is largely inspired by the work carried out in NMT and MNMT. Component such as attention models like multi-head, encoder-based and pooled attention (\citeyear{W17-4746,DBLP:journals/corr/abs-1712-03449,NIPS2017_7181}); reinforcement learning in NMT and MNMT (\citeyear{P16-1159,emnlpjb}); embeddings (\citeyear{P16-1162,Delbrouck2017visually}) are well investigated.\\

In captioning, \citeauthor{Anderson2017up-down} used a very similar approach where two LSTMs build and encode the visual features. Two other works, \citeauthor{yao2017boosting} and \citeauthor{You2016ImageCW} used pooled features as described in this paper. However, they both used an additional vector taken from the fully connected layer of a CNN.

\section{Conclusion}
We presented a novel and light architecture composed of a cGRU that showed interesting performance. The model builds and encodes the context vector from pooled features in an efficient manner. The attention model presented in section \ref{attmodel} is really straightforward and seems to bring the necessary visual information in order to output complete captions. Also, we empirically showed that the model can easily scale with more sought-after modules or simple with more parameters. In the future, it would be interesting to use different attention features, like VGG or GoogleNet (that have only 1024 dimensions) or different attention models to see how far this architecture can get.

\section{Acknowledgements}
This work was partly supported by the Chist-Era project IGLU with contribution from the Belgian Fonds de la Recherche Scientique (FNRS), contract no. R.50.11.15.F, and by the FSO project VCYCLE with contribution from the Belgian Waloon Region, contract no. 1510501. \\

We also thank the authors of nmtpytorch\footnote{https://github.com/lium-lst/nmtpytorch} \cite{nmtpy2017} that we used as framework for our experiments. Our code is made available for posterior research \footnote{https://github.com/jbdel/light\_captioning}.

\newpage

\bibliographystyle{aaai19}
\bibliography{biblio.bib}

\begin{thebibliography}{}

\bibitem[\protect\citeauthoryear{Anderson \bgroup et al\mbox.\egroup
  }{2018}]{Anderson2017up-down}
Anderson, P.; He, X.; Buehler, C.; Teney, D.; Johnson, M.; Gould, S.; and
  Zhang, L.
\newblock 2018.
\newblock Bottom-up and top-down attention for image captioning and visual
  question answering.
\newblock In {\em CVPR}.

\bibitem[\protect\citeauthoryear{Bahdanau, Cho, and
  Bengio}{2014}]{bahdanau+al-2014-nmt}
Bahdanau, D.; Cho, K.; and Bengio, Y.
\newblock 2014.
\newblock Neural machine translation by jointly learning to align and
  translate.
\newblock {\em arXiv e-prints} abs/1409.0473.

\bibitem[\protect\citeauthoryear{Caglayan \bgroup et al\mbox.\egroup
  }{2017a}]{W17-4746}
Caglayan, O.; Aransa, W.; Bardet, A.; Garc{\'i}a-Mart{\'i}nez, M.; Bougares,
  F.; Barrault, L.; Masana, M.; Herranz, L.; and van~de Weijer, J.
\newblock 2017a.
\newblock Lium-cvc submissions for wmt17 multimodal translation task.
\newblock In {\em Proceedings of the Second Conference on Machine Translation},
   432--439.
\newblock Association for Computational Linguistics.

\bibitem[\protect\citeauthoryear{Caglayan \bgroup et al\mbox.\egroup
  }{2017b}]{nmtpy2017}
Caglayan, O.; Garc\'{i}a-Mart\'{i}nez, M.; Bardet, A.; Aransa, W.; Bougares,
  F.; and Barrault, L.
\newblock 2017b.
\newblock Nmtpy: A flexible toolkit for advanced neural machine translation
  systems.
\newblock {\em Prague Bull. Math. Linguistics} 109:15--28.

\bibitem[\protect\citeauthoryear{Chen \bgroup et al\mbox.\egroup
  }{2015}]{DBLP:journals/corr/ChenFLVGDZ15}
Chen, X.; Fang, H.; Lin, T.; Vedantam, R.; Gupta, S.; Doll{\'{a}}r, P.; and
  Zitnick, C.~L.
\newblock 2015.
\newblock Microsoft {COCO} captions: Data collection and evaluation server.
\newblock {\em CoRR} abs/1504.00325.

\bibitem[\protect\citeauthoryear{Cho \bgroup et al\mbox.\egroup
  }{2014}]{cho-al-emnlp14}
Cho, K.; van Merri{\"{e}}nboer, B.; G{\"{u}}l{\c c}ehre, {\c C}.; Bahdanau, D.;
  Bougares, F.; Schwenk, H.; and Bengio, Y.
\newblock 2014.
\newblock Learning phrase representations using rnn encoder--decoder for
  statistical machine translation.
\newblock In {\em Proceedings of the 2014 Conference on Empirical Methods in
  Natural Language Processing (EMNLP)},  1724--1734.
\newblock Doha, Qatar: Association for Computational Linguistics.

\bibitem[\protect\citeauthoryear{Delbrouck and Dupont}{2017a}]{emnlpjb}
Delbrouck, J.-B., and Dupont, S.
\newblock 2017a.
\newblock An empirical study on the effectiveness of images in multimodal
  neural machine translation.
\newblock In {\em Proceedings of the 2017 Conference on Empirical Methods in
  Natural Language Processing},  910--919.
\newblock Association for Computational Linguistics.

\bibitem[\protect\citeauthoryear{Delbrouck and
  Dupont}{2017b}]{DBLP:journals/corr/abs-1712-03449}
Delbrouck, J., and Dupont, S.
\newblock 2017b.
\newblock Modulating and attending the source image during encoding improves
  multimodal translation.
\newblock {\em CoRR} abs/1712.03449.

\bibitem[\protect\citeauthoryear{Delbrouck and
  Dupont}{2018}]{delbrouck2018umons}
Delbrouck, J.-B., and Dupont, S.
\newblock 2018.
\newblock Umons submission for wmt18 multimodal translation task.
\newblock In {\em Proceedings of the First Conference on Machine Translation}.
\newblock Brussels, Belgium: Association for Computational Linguistics.

\bibitem[\protect\citeauthoryear{Delbrouck, Dupont, and
  Seddati}{2017}]{Delbrouck2017visually}
Delbrouck, J.-B.; Dupont, S.; and Seddati, O.
\newblock 2017.
\newblock Visually grounded word embeddings and richer visual features for
  improving multimodal neural machine translation.
\newblock In {\em Proc. GLU 2017 International Workshop on Grounding Language
  Understanding},  62--67.

\bibitem[\protect\citeauthoryear{He \bgroup et al\mbox.\egroup
  }{2016}]{He2016DeepRL}
He, K.; Zhang, X.; Ren, S.; and Sun, J.
\newblock 2016.
\newblock Deep residual learning for image recognition.
\newblock {\em 2016 IEEE Conference on Computer Vision and Pattern Recognition
  (CVPR)}  770--778.

\bibitem[\protect\citeauthoryear{Hochreiter and
  Schmidhuber}{1997}]{Hochreiter:1997:LSM:1246443.1246450}
Hochreiter, S., and Schmidhuber, J.
\newblock 1997.
\newblock Long short-term memory.
\newblock {\em Neural Comput.} 9(8):1735--1780.

\bibitem[\protect\citeauthoryear{Karpathy and Li}{2015}]{conf/cvpr/KarpathyL15}
Karpathy, A., and Li, F.-F.
\newblock 2015.
\newblock Deep visual-semantic alignments for generating image descriptions.
\newblock In {\em CVPR},  3128--3137.
\newblock IEEE Computer Society.

\bibitem[\protect\citeauthoryear{Kingma and Ba}{2014}]{kingma2014adam}
Kingma, D.~P., and Ba, J.
\newblock 2014.
\newblock Adam: A method for stochastic optimization.
\newblock {\em arXiv preprint arXiv:1412.6980}.

\bibitem[\protect\citeauthoryear{Lavie and
  Agarwal}{2007}]{Lavie:2007:MAM:1626355.1626389}
Lavie, A., and Agarwal, A.
\newblock 2007.
\newblock Meteor: An automatic metric for mt evaluation with high levels of
  correlation with human judgments.
\newblock In {\em Proceedings of the Second Workshop on Statistical Machine
  Translation}, StatMT '07,  228--231.
\newblock Stroudsburg, PA, USA: Association for Computational Linguistics.

\bibitem[\protect\citeauthoryear{Lin \bgroup et al\mbox.\egroup
  }{2014}]{mscoco}
Lin, T.-Y.; Maire, M.; Belongie, S.; Hays, J.; Perona, P.; Ramanan, D.;
  Doll{\'a}r, P.; and Zitnick, C.~L.
\newblock 2014.
\newblock Microsoft coco: Common objects in context.
\newblock In Fleet, D.; Pajdla, T.; Schiele, B.; and Tuytelaars, T., eds., {\em
  Computer Vision -- ECCV 2014},  740--755.
\newblock Cham: Springer International Publishing.

\bibitem[\protect\citeauthoryear{Liu \bgroup et al\mbox.\egroup
  }{2017}]{Liu2017ImprovedIC}
Liu, S.; Zhu, Z.; Ye, N.; Guadarrama, S.; and Murphy, K.
\newblock 2017.
\newblock Improved image captioning via policy gradient optimization of spider.
\newblock {\em 2017 IEEE International Conference on Computer Vision (ICCV)}
  873--881.

\bibitem[\protect\citeauthoryear{Lu \bgroup et al\mbox.\egroup
  }{2017}]{Lu2017Adaptive}
Lu, J.; Xiong, C.; Parikh, D.; and Socher, R.
\newblock 2017.
\newblock Knowing when to look: Adaptive attention via a visual sentinel for
  image captioning.

\bibitem[\protect\citeauthoryear{Mikolov \bgroup et al\mbox.\egroup
  }{2010}]{conf/interspeech/MikolovKBCK10}
Mikolov, T.; Karafiát, M.; Burget, L.; Cernocký, J.; and Khudanpur, S.
\newblock 2010.
\newblock Recurrent neural network based language model.
\newblock In Kobayashi, T.; Hirose, K.; and Nakamura, S., eds., {\em
  INTERSPEECH},  1045--1048.
\newblock ISCA.

\bibitem[\protect\citeauthoryear{Mikolov \bgroup et al\mbox.\egroup
  }{2013}]{NIPS2013_5021}
Mikolov, T.; Sutskever, I.; Chen, K.; Corrado, G.~S.; and Dean, J.
\newblock 2013.
\newblock Distributed representations of words and phrases and their
  compositionality.
\newblock In Burges, C. J.~C.; Bottou, L.; Welling, M.; Ghahramani, Z.; and
  Weinberger, K.~Q., eds., {\em Advances in Neural Information Processing
  Systems 26}. Curran Associates, Inc.
\newblock  3111--3119.

\bibitem[\protect\citeauthoryear{Mun, Cho, and Han}{2017}]{mun2017textguided}
Mun, J.; Cho, M.; and Han, B.
\newblock 2017.
\newblock Text-guided attention model for image captioning.
\newblock In {\em AAAI}.

\bibitem[\protect\citeauthoryear{Papineni \bgroup et al\mbox.\egroup
  }{2002}]{Papineni:2002:BMA:1073083.1073135}
Papineni, K.; Roukos, S.; Ward, T.; and Zhu, W.-J.
\newblock 2002.
\newblock Bleu: A method for automatic evaluation of machine translation.
\newblock In {\em Proceedings of the 40th Annual Meeting on Association for
  Computational Linguistics}, ACL '02,  311--318.
\newblock Stroudsburg, PA, USA: Association for Computational Linguistics.

\bibitem[\protect\citeauthoryear{Pennington, Socher, and
  Manning}{2014}]{pennington2014glove}
Pennington, J.; Socher, R.; and Manning, C.~D.
\newblock 2014.
\newblock Glove: Global vectors for word representation.
\newblock In {\em EMNLP}, volume~14,  1532--1543.

\bibitem[\protect\citeauthoryear{Ranzato \bgroup et al\mbox.\egroup
  }{2015}]{DBLP:journals/corr/RanzatoCAZ15}
Ranzato, M.; Chopra, S.; Auli, M.; and Zaremba, W.
\newblock 2015.
\newblock Sequence level training with recurrent neural networks.
\newblock {\em CoRR} abs/1511.06732.

\bibitem[\protect\citeauthoryear{Rennie \bgroup et al\mbox.\egroup
  }{2017}]{Rennie2017SelfCriticalST}
Rennie, S.~J.; Marcheret, E.; Mroueh, Y.; Ross, J.; and Goel, V.
\newblock 2017.
\newblock Self-critical sequence training for image captioning.
\newblock {\em 2017 IEEE Conference on Computer Vision and Pattern Recognition
  (CVPR)}  1179--1195.

\bibitem[\protect\citeauthoryear{Sennrich, Haddow, and Birch}{2016}]{P16-1162}
Sennrich, R.; Haddow, B.; and Birch, A.
\newblock 2016.
\newblock Neural machine translation of rare words with subword units.
\newblock In {\em Proceedings of the 54th Annual Meeting of the Association for
  Computational Linguistics (Volume 1: Long Papers)},  1715--1725.
\newblock Association for Computational Linguistics.

\bibitem[\protect\citeauthoryear{Shen \bgroup et al\mbox.\egroup
  }{2016}]{P16-1159}
Shen, S.; Cheng, Y.; He, Z.; He, W.; Wu, H.; Sun, M.; and Liu, Y.
\newblock 2016.
\newblock Minimum risk training for neural machine translation.
\newblock In {\em Proceedings of the 54th Annual Meeting of the Association for
  Computational Linguistics (Volume 1: Long Papers)},  1683--1692.
\newblock Association for Computational Linguistics.

\bibitem[\protect\citeauthoryear{Simonyan and
  Zisserman}{2014}]{Simonyan2014VeryDC}
Simonyan, K., and Zisserman, A.
\newblock 2014.
\newblock Very deep convolutional networks for large-scale image recognition.
\newblock {\em CoRR} abs/1409.1556.

\bibitem[\protect\citeauthoryear{Specia \bgroup et al\mbox.\egroup
  }{2016}]{specia-EtAl:2016:WMT}
Specia, L.; Frank, S.; Sima'an, K.; and Elliott, D.
\newblock 2016.
\newblock A shared task on multimodal machine translation and crosslingual
  image description.
\newblock In {\em Proceedings of the First Conference on Machine Translation},
  543--553.
\newblock Berlin, Germany: Association for Computational Linguistics.

\bibitem[\protect\citeauthoryear{Sutskever, Vinyals, and
  Le}{2014}]{SutskeverVL14}
Sutskever, I.; Vinyals, O.; and Le, Q.~V.
\newblock 2014.
\newblock Sequence to sequence learning with neural networks.
\newblock In {\em Advances in neural information processing systems},
  3104--3112.

\bibitem[\protect\citeauthoryear{Vaswani \bgroup et al\mbox.\egroup
  }{2017}]{NIPS2017_7181}
Vaswani, A.; Shazeer, N.; Parmar, N.; Uszkoreit, J.; Jones, L.; Gomez, A.~N.;
  Kaiser, L.~u.; and Polosukhin, I.
\newblock 2017.
\newblock Attention is all you need.
\newblock In Guyon, I.; Luxburg, U.~V.; Bengio, S.; Wallach, H.; Fergus, R.;
  Vishwanathan, S.; and Garnett, R., eds., {\em Advances in Neural Information
  Processing Systems 30}. Curran Associates, Inc.
\newblock  5998--6008.

\bibitem[\protect\citeauthoryear{Vedantam, Lawrence~Zitnick, and
  Parikh}{2015}]{Vedantam_2015_CVPR}
Vedantam, R.; Lawrence~Zitnick, C.; and Parikh, D.
\newblock 2015.
\newblock Cider: Consensus-based image description evaluation.
\newblock In {\em The IEEE Conference on Computer Vision and Pattern
  Recognition (CVPR)}.

\bibitem[\protect\citeauthoryear{Wu \bgroup et al\mbox.\egroup
  }{2016}]{Wu_2016_CVPR}
Wu, Q.; Shen, C.; Liu, L.; Dick, A.; and van~den Hengel, A.
\newblock 2016.
\newblock What value do explicit high level concepts have in vision to language
  problems?
\newblock In {\em The IEEE Conference on Computer Vision and Pattern
  Recognition (CVPR)}.

\bibitem[\protect\citeauthoryear{Xu \bgroup et al\mbox.\egroup
  }{2015}]{pmlr-v37-xuc15}
Xu, K.; Ba, J.; Kiros, R.; Cho, K.; Courville, A.; Salakhudinov, R.; Zemel, R.;
  and Bengio, Y.
\newblock 2015.
\newblock Show, attend and tell: Neural image caption generation with visual
  attention.
\newblock In Bach, F., and Blei, D., eds., {\em Proceedings of the 32nd
  International Conference on Machine Learning}, volume~37 of {\em Proceedings
  of Machine Learning Research},  2048--2057.
\newblock Lille, France: PMLR.

\bibitem[\protect\citeauthoryear{Yang \bgroup et al\mbox.\egroup
  }{2016}]{YangReview}
Yang, Z.; Yuan, Y.; Wu, Y.; Cohen, W.~W.; and Salakhutdinov, R.~R.
\newblock 2016.
\newblock Review networks for caption generation.
\newblock In Lee, D.~D.; Sugiyama, M.; Luxburg, U.~V.; Guyon, I.; and Garnett,
  R., eds., {\em Advances in Neural Information Processing Systems 29}. Curran
  Associates, Inc.
\newblock  2361--2369.

\bibitem[\protect\citeauthoryear{Yao \bgroup et al\mbox.\egroup
  }{2017}]{yao2017boosting}
Yao, T.; Pan, Y.; Li, Y.; Qiu, Z.; and Mei, T.
\newblock 2017.
\newblock Boosting image captioning with attributes.
\newblock In {\em ICCV}.

\bibitem[\protect\citeauthoryear{You \bgroup et al\mbox.\egroup
  }{2016}]{You2016ImageCW}
You, Q.; Jin, H.; Wang, Z.; Fang, C.; and Luo, J.
\newblock 2016.
\newblock Image captioning with semantic attention.
\newblock {\em 2016 IEEE Conference on Computer Vision and Pattern Recognition
  (CVPR)}  4651--4659.

\end{thebibliography}
\end{document}